\documentclass[AMA,Times2COL]{WileyNJDv5} 

\articletype{Article Type}%

\startpage{1}

\usepackage{hyperref}
\usepackage{graphics} 
\usepackage{epsfig} 
\usepackage{mathptmx} 
\usepackage{times} 
\usepackage{amsmath} 
\usepackage{amssymb}  
\usepackage{graphicx}
\usepackage{subfigure}
\usepackage{multirow}
\usepackage{bm}
\usepackage{makecell}

\begin{document}

\title{SPF-EMPC Planner: A real-time multi-robot trajectory planner for complex environments with uncertainties}

\author{Peng Liu}
\author{Pengming Zhu}
\author{Zhiwen Zeng}
\author{Xuekai Qiu}
\author{Yu Wang}
\author{Huimin Lu}

\authormark{Liu \textsc{et al.}}
\titlemark{SPF-EMPC Planner: A real-time multi-robot trajectory planner for complex environments with uncertainties}

\address{\orgdiv{College of Intelligence Science and Technology}, \orgname{National University of Defense Technology}, \orgaddress{\state{Changsha}, \country{China}}}

\corres{Corresponding author Zhiwen Zeng. \email{zengzhiwen@nudt.edu.cn}}


\fundingInfo{National Science Foundation of China, Grant/Award Number: 62203460, U22A2059; Major Project of the Natural Science Foundation of Hunan Province, Grant/Award Number: 2021JC0004}

\abstract[Abstract]{In practical applications, the unpredictable movement of obstacles and the imprecise state observation of robots introduce significant uncertainties for the swarm of robots, especially in cluster environments. However, existing methods are difficult to realize safe navigation, considering uncertainties, complex environmental structures, and robot swarms. This paper introduces an extended state model predictive control planner with a safe probability field to address the multi-robot navigation problem in complex, dynamic, and uncertain environments. Initially, the safe probability field offers an innovative approach to model the uncertainty of external dynamic obstacles, combining it with an unconstrained optimization method to generate safe trajectories for multi-robot online. Subsequently, the extended state model predictive controller can accurately track these generated trajectories while considering the robots' inherent model constraints and state uncertainty, thus ensuring the practical feasibility of the planned trajectories. Simulation experiments show a success rate four times higher than that of state-of-the-art algorithms. Physical experiments demonstrate the method's ability to operate in real-time, enabling safe navigation for multi-robot in uncertain environments.}



\maketitle


\renewcommand\thefootnote{\fnsymbol{footnote}}
\setcounter{footnote}{1}

\section{INTRODUCTION}\label{sec1}

\begin{figure}[htbp]
	\centering
	\includegraphics[width = 1.0\linewidth, keepaspectratio]{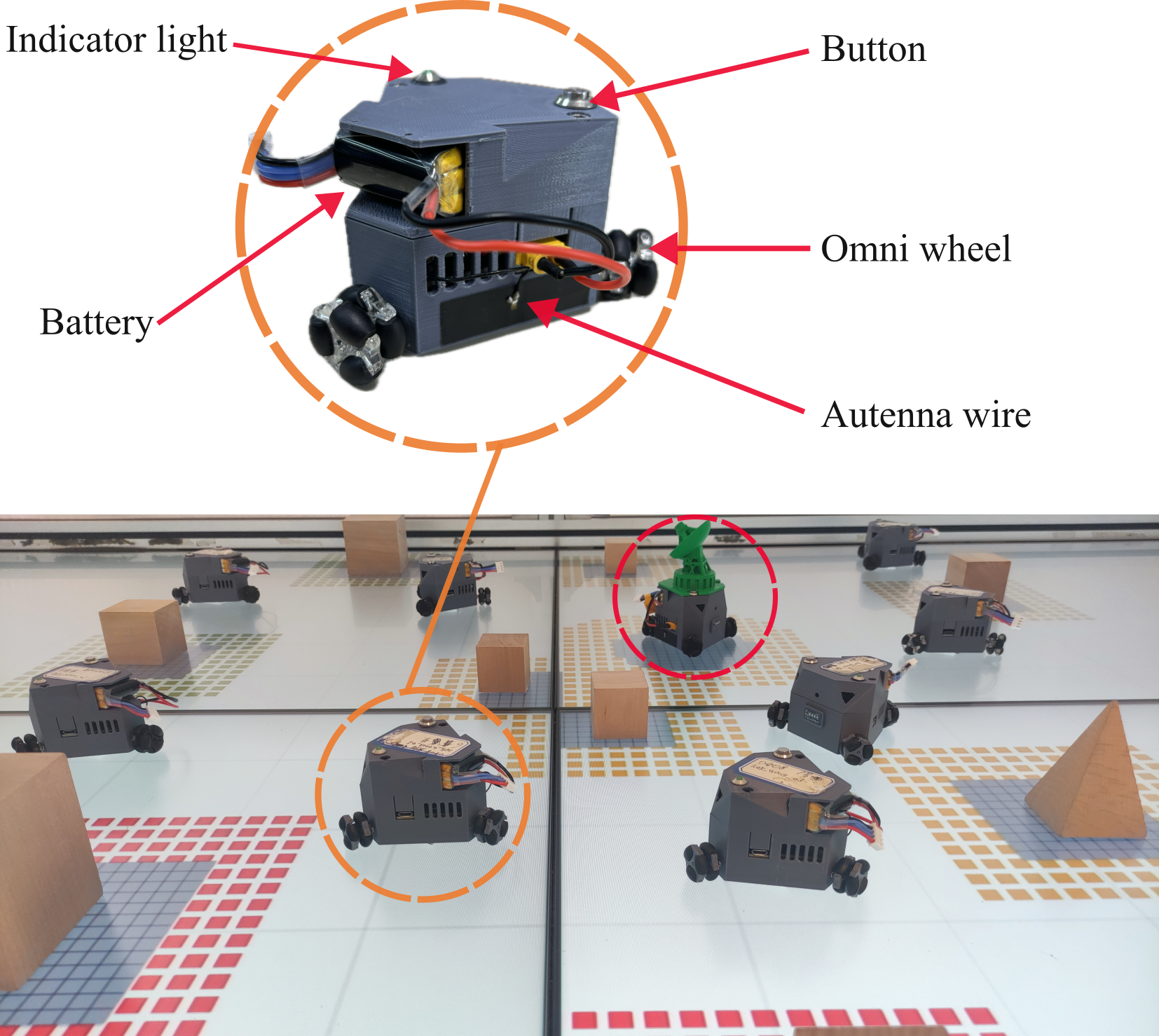}
	\caption{The ground mobile robots' navigation in a complex environment using the proposed algorithm. The upper shows the three-wheeled omnidirectional mobile robot used for the real-world experiment. The bottom presents the navigation in a physical environment with static, moving obstacles (the one in the red circle marked in green) and other robots. }
	\label{fig1}
\end{figure}

Multi-robot systems are one of the most widely researched areas in robotics. For these robots, safe and fast cooperative navigation is the key to accomplishing their tasks. However, in a complex environment containing static obstacles, dynamic obstacles, and multiple robots, various factors may introduce uncertainties that greatly affect the planning and control of multi-robot systems. As shown in Fig. \ref{fig1}, robots are required to safely navigate to their destinations in a cluttered environment, where potential collision risks emerge from unpredictable dynamic obstacles and the robot's inaccurate state observation. Consequently, adopting a safe trajectory planning framework in complex environments that account for uncertainties is essential.

Two main challenges in current multi-robot planning make the problem studied in this paper not effectively addressed in previous research: Firstly, previous work rely on specific maps such as occupy maps \cite{hornung2013octomap} and ESDF maps \cite{han2019fiesta}, which are unable to represent dynamic obstacles adequately and lack the necessary consideration of uncertain environment. Secondly, it is challenging to achieve scalability and robustness, considering complex environments, multiple robots, and control biases in real-world applications. While some work like ego-serials \cite{zhou2020ego,zhou2021ego,zhou2022swarm} can navigate robots in static environments, they do not consider the additional characteristics of dynamic obstacles, making it difficult to avoid them. Other work \cite{lin2020robust,castillo2020real} employ geometric representations and model predictive control to avoid dynamic obstacles but overlook the complex static environment structures. On the whole, few studies have considered the uncertainties caused by the unpredictable movement of obstacles and the imprecise observation of states in multi-robot systems.

To address the uncertainties mentioned above, this paper proposes a novel unified framework for multi-robot trajectory planning and tracking: Safety Probability Field based Extended state Model Predictive Control planner (SPF-EMPC), in which a light local trajectory planner is incorporated in an iterative trajectory tracker. In trajectory planning, a safety probability field is constructed to model the uncertainties associated with obstacles. Requirements such as obstacle avoidance and smoothness are integrated into an efficient unconstrained optimization for trajectory generation, thereby reducing the computational load. For trajectory tracking, the state uncertainty and the model constraints are considered. The model predictive control employed with an extended state observer and a temporal self-sampling method enables the robot to track the temporal target accurately, even in the face of state uncertainty, ensuring the effectiveness of the planned trajectory. The overall framework achieves significant advantages in terms of success rate and number of iterations.

This paper assumes that the uncertainties of the robot state and dynamic obstacles are bounded. Additionally, the dynamic obstacles have a maximum velocity limit and are more likely to appear in the direction of the current speed. The main contributions are summarized as follows:
\begin{itemize}
\item[$\bullet$] \textbf{Online trajectory generator for uncertain obstacles:} We construct a safe probability field based on Gaussian distribution for dynamic obstacles with uncertainty and apply an unconstrained optimization method to generate safe trajectories for multi-robot online in complex environments.
\item[$\bullet$] \textbf{Trajectory tracker based on extended state model predictive control:} We design extended state model predictive control to handle the uncertainty of the robot's observed state and accurately track the trajectory through a temporal self-sampling approach.
\item[$\bullet$] \textbf{Simulation and physical validation:} We conducted multi-robot traversal experiments in both simulation and physical environments to validate the proposed method's real-time performance and effectiveness.
\end{itemize}
\section{Related work}

In recent years, trajectory planning in complex static environments using voxel maps \cite{hornung2013octomap,han2019fiesta} and optimization methods have made significant progress in simulation and experiment. For instance, Xu \cite{csenbacslar2023rlss} employs the safe flight corridor approach, which utilizes the obstacle space as a constraint to guarantee collision-free trajectories, but the computational cost is expensive when executed online. Then, inspired by the work \cite{ratliff2009chomp,mellinger2011minimum,mellinger2012mixed}, gradient-based planner \cite{zhou2019robust} employs the differential flatness property of quadrotor unmanned aerial vehicles (UAVs) and the distance information of ESDF maps to simplify and unconstrained transform the optimization problem. Later, EGO \cite{zhou2020ego} eliminates the dependence on ESDF maps, reduces the computation, and has been successfully extended to several versions \cite{zhou2021ego,zhou2022swarm,xu2023vision}. Although the above methods have demonstrated high computational efficiency and success rate in navigation tests, the voxel maps they rely on can not adequately describe dynamic obstacles with uncertainties, resulting in limited performance in dynamic and uncertain environments.

\begin{table*}[ht]
\caption{Comparison of recent motion planning methods according to key attributes in the table.}
\centering
\begin{tabular}{ccccc}
\hline
Method         & Group Scale                  & Static Structure              & Uncertain Consideration               & Verification Method               \\ \hline
EGO-Swarm\cite{zhou2021ego}      & {\color[HTML]{009901} swarm}                         & {\color[HTML]{009901} complex}                       & {\color[HTML]{FF0000} not considered} & {\color[HTML]{009901} hardware}                          \\
EGO-Swarm2\cite{zhou2022swarm}     & {\color[HTML]{009901} swarm}                        & {\color[HTML]{009901} complex}                        & {\color[HTML]{FF0000} not considered} & {\color[HTML]{009901} hardware}                          \\
MADER\cite{tordesillas2021mader}          & {\color[HTML]{009901} swarm}                         & {\color[HTML]{009901} complex}                        & {\color[HTML]{FF0000} not considered} & {\color[HTML]{FF0000} simulation} \\
RMADER\cite{kondo2023robust}         & {\color[HTML]{009901} swarm}                         & {\color[HTML]{FF0000} simple} & {\color[HTML]{FF0000} not considered} & {\color[HTML]{009901} hardware}                          \\
DPMPC\cite{xu2022dpmpc}          & {\color[HTML]{FF0000} single} & {\color[HTML]{009901} complex}                        & {\color[HTML]{009901} obstacles}                              & {\color[HTML]{FF0000} simulation} \\
VIGO\cite{xu2023vision}           & {\color[HTML]{FF0000} single} & {\color[HTML]{009901} complex}                        & {\color[HTML]{FF0000} not considered} & {\color[HTML]{009901} hardware}                          \\
CCNMPC\cite{zhu2019chance}         & {\color[HTML]{009901} swarm}                        & {\color[HTML]{FF0000} simple} & {\color[HTML]{009901} obstacles}                             & {\color[HTML]{009901} hardware}                          \\
SPF-EMPC(Ours) & {\color[HTML]{009901} swarm}                         & {\color[HTML]{009901} complex}                        & {\color[HTML]{009901} obstacles and robot states}            & {\color[HTML]{009901} hardware}                          \\ \hline
\end{tabular}
\label{tab1}
\end{table*}

For dynamic obstacles, early methods based on geometric maps proposed artificial potential field \cite{khatib1986real} and velocity obstacle \cite{guo2021vr} to generate simple control commands to prevent collisions. Recently, model predictive control (MPC) approaches \cite{lin2020robust,zhu2019chance,guo2022obstacle} have emerged as dynamic obstacle avoidance methods. MPC can be integrated with dynamic obstacle avoidance to generate and track trajectories that satisfy kinematic, dynamic, and actuation constraints. In the work\cite{xu2022dpmpc}, obstacles are wrapped as ellipses, and a distance cost is imposed as a trajectory penalty. To further consider the uncertainty of obstacles, Jian\cite{jian2023dynamic} integrates a control barrier function within the MPC framework to avoid pedestrians. Meanwhile, some work\cite{lin2020robust,zhu2019chance} reduce the computation by linearizing approximate chance constraints, thereby taking uncertainty into account and realizing real-time obstacle avoidance. These methods represent obstacles in terms of geometric shapes, which may lead to collisions and suboptimal performance when facing complex static structures. Moreover, it is challenging to scale up to multi-robot navigation in cluttered environments, given the state uncertainty and the computational consumption of open-source MPC libraries \cite{andersson2019casadi,houska2011acado}.

A few methods employ a two-layer framework to address static and dynamic obstacles separately. Xu\cite{xu2022dpmpc} utilizes voxel map-based point cloud detection with a linear prediction of obstacle states for safe navigation. However, the linear prediction approach for forecasting the future states of obstacles may be overly conservative. Furthermore, this framework's global static trajectory remains unchanged, potentially failing to find solutions that account for diverse environmental uncertainties.

In this paper, beyond the special uncertain environment, the studied ground mobile robots have limited operational ranges and harder obstacle avoidance than UAVs. As shown in Table \ref{tab1}, unlike previous work, this paper considers uncertainties and conducts physical verification in complex environments. The proposed method separately models and filters various uncertainties and leverages the effectiveness of unconstrained optimization methods and the robustness of model predictive control to generate and track safe trajectories online for multi-robot in uncertain environments.

\section{Methodology}\label{sec2}

\subsection{System Framework Overview}
\begin{figure}[htbp]
	\centering
	\includegraphics[width = 1.0\linewidth, keepaspectratio]{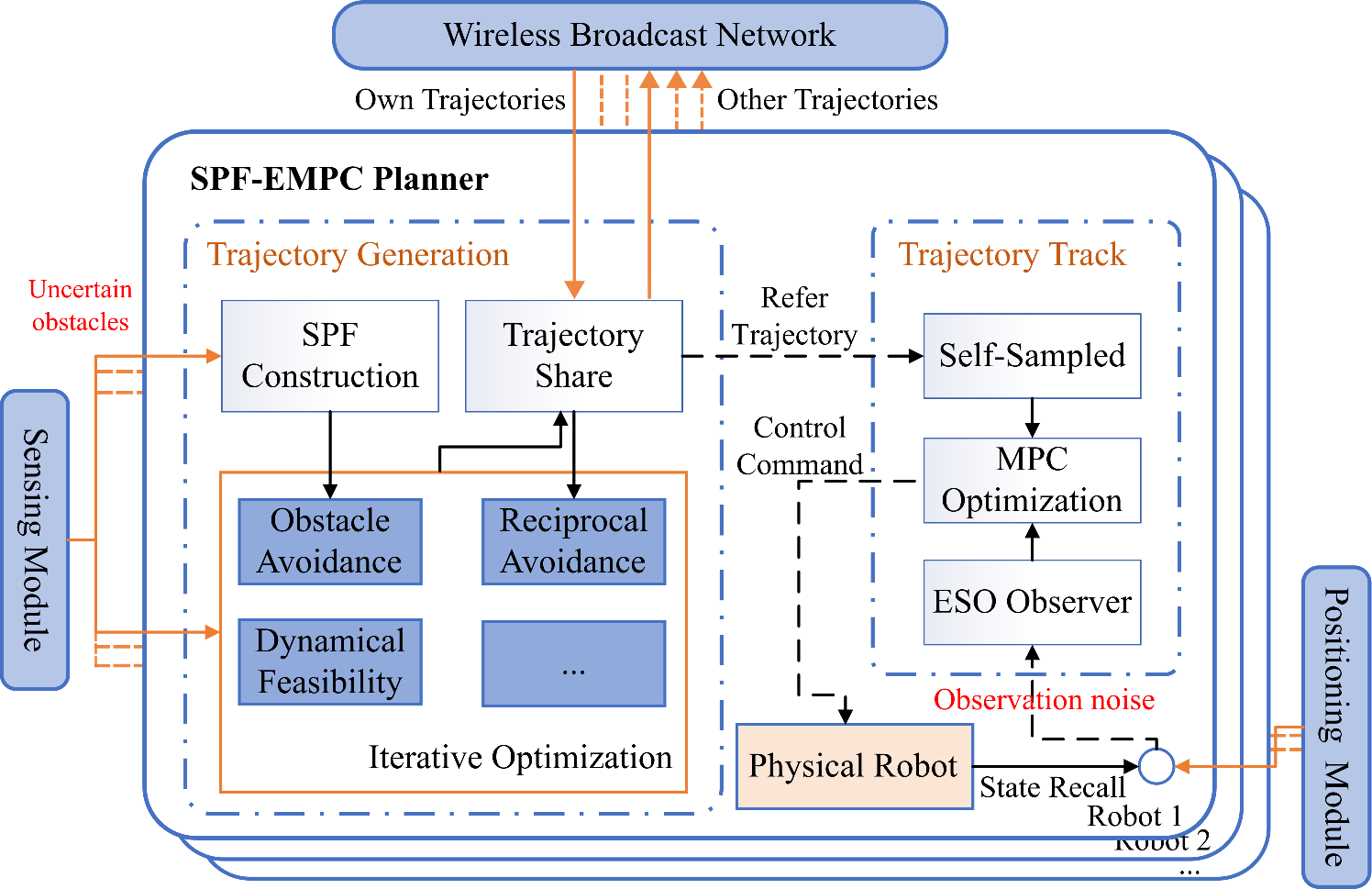}
	\caption{System overview. When perceiving the dynamic obstacle, a probability field is generated in real-time based on the obstacle's state. Then, the local trajectory planner iteratively generates the safe trajectory based on the local environment and probability field. Finally, the trajectory with the requisite time is tracked by the extended state model predictive controller. Throughout this process, the perception and localization modules regularly provide collision detection for each robot.}
	\label{fig2}
\end{figure}

This work introduces a multi-robot distributed trajectory planning framework encompassing three principal components: the sensing and position module, the local trajectory planning module, and the extended state model predictive control module, as shown in Fig. \ref{fig2}.

Each robot has a distinct sensing range within this distributed system for its planning and control processes. And trajectories are communicated through a prioritization mechanism\footnote{This paper sets a fixed priority according to the robot number. Robots with a lower priority need to consider obstacle avoidance with other robots with a higher priority.}.

\subsection{Safety Probability Field}
In contrast to the fixed nature of static obstacles, dynamic obstacles may appear anywhere within the vicinity, introducing uncertainty. This section describes the construction of the safety probability field for uncertain obstacles. 

\begin{figure}[htbp]
	\centering
	\includegraphics[width = 0.8\linewidth, keepaspectratio]{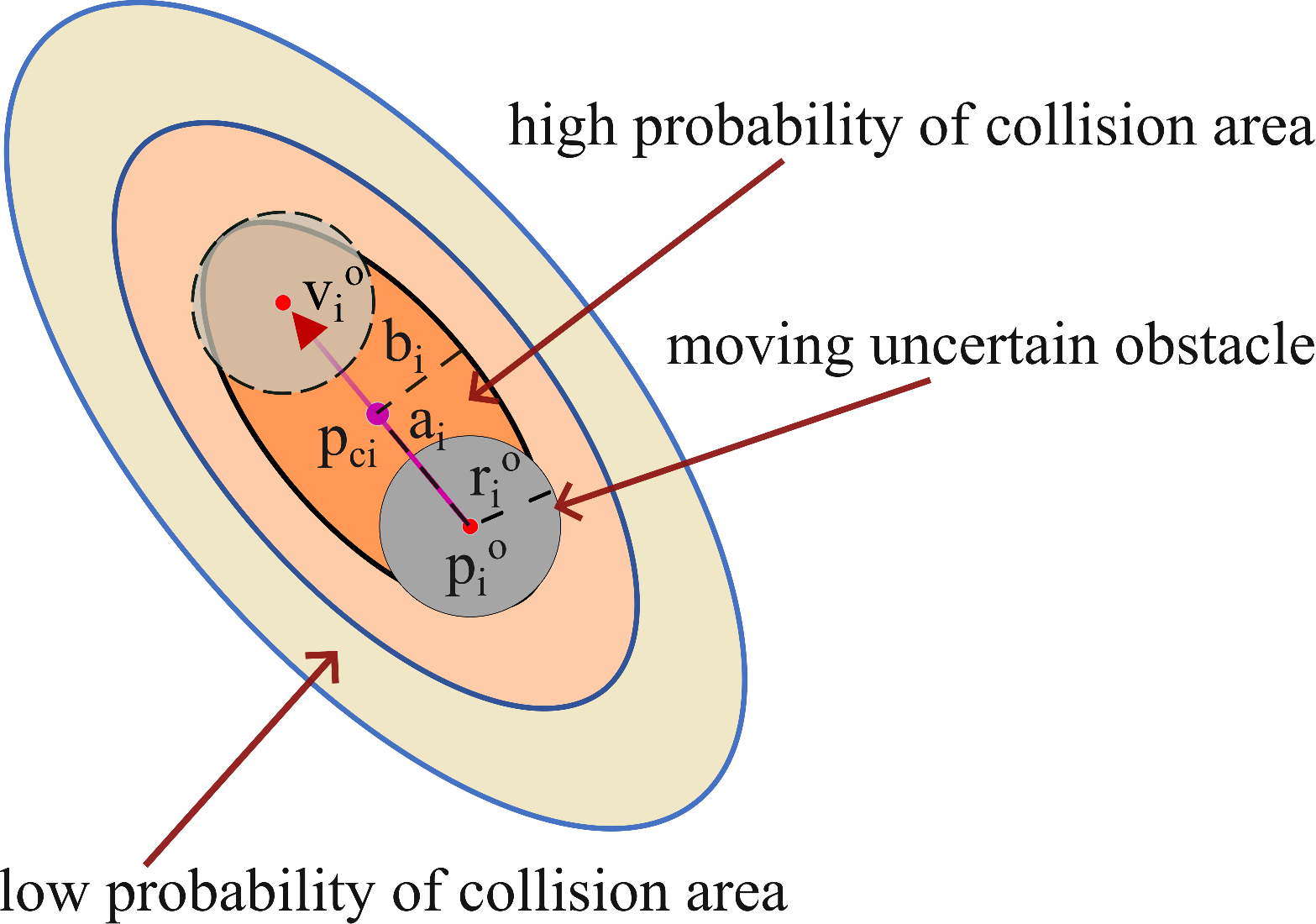}
	\caption{Illustration of the safety probability field. The field is generated based on the dynamic obstacle's current shape, position, and velocity, with different collision probabilities for different regions. The transparent circle surrounded by a dashed line represents the state of the obstacle at the next moment during deterministic motion.}
	\label{fig3}
\end{figure}

This work enhances the real-time performance and effectiveness of obstacle avoidance by referring to studies on pedestrian prediction \cite{bhatt2023mpc,zhou2022human} and introducing an elliptic dynamic probability field with a binary Gaussian distribution, as shown in Figure \ref{fig3}. Compared to a general circular field, this safety probability field can describe the moving intention of obstacles, giving the robot precise guidance on movement. Safe trajectories that consider finite future states and avoid being overly conservative are generated by avoiding regions with a high collision probability.

The probability field is designed concerning the Gaussian distribution as follows:
\begin{equation}
	\begin{aligned}
	U\left( {{\bf{p}}_i^o,{\bf{v}}_i^o,{\bf{p}}} \right) = \frac{{\exp ( - \frac{1}{2}{{({\bf{p}} - {\bf{p}}_{{c_i}}^o)}^{\mathrm{T}}}{{\bm{\Sigma }}^{ - 1}}({\bf{p}} - {\bf{p}}_{{c_i}}^o))}}{{2{\rm{\pi }}|{\bm{\Sigma }}{|^{1/2}}}}\,,
		\label{eq1}
	\end{aligned}
\end{equation}
\begin{equation}
	\mathbf{p}_{{{c}_{i}}}^{o}=\mathbf{p}_{i}^{o}+\left(1-\lambda\right) \mathbf{v}_{i}^{o}dt,\lambda \in [0,1]\,,
	\label{eq2}
\end{equation}
where ${\bf{p}}_i^o$ and ${\bf{v}}_i^o$ represent the current position and velocity of the $i{\rm{th}}$ obstacle, and ${\bf{p}}$ denotes the position to be queried. The ${\bf{p}}_{{c_i}}^o$ corresponds to the central position of the probability field, $dt$ is the sampling interval, and $\lambda$ indicates the degree of attention to the current position. The function value $U$ means the likelihood of a dynamic obstacle's presence at location ${\bf{p}}$.

For the covariance matrix ${\bm{\Sigma}}$, we set $\theta_i  = \arctan \left( {{{{\bf{v}}_{iy}^o} \mathord{\left/
{\vphantom {{{\bf{v}}_{iy}^o} {{\bf{v}}_{ix}^o}}} \right.
\kern-\nulldelimiterspace} {{\bf{v}}_{ix}^o}}} \right)$ so that the rotation matrix $\bf{R}$ and deflation matrix $\bm{\Lambda }$ are related to the obstacle's size and velocity:
\begin{equation}
	\begin{array}{c}
\Sigma  = {\bf{R}}\Lambda {{\bf{R}}^{\rm{T}}}\\
 = \left[ {\begin{array}{*{20}{c}}
{\cos {\theta _i}}&{ - \sin {\theta _i}}\\
{\sin {\theta _i}}&{\cos {\theta _i}}
\end{array}} \right]\left[ {\begin{array}{*{20}{c}}
{{a_i}^2}&0\\
0&{{b_i}^2}
\end{array}} \right]{\left[ {\begin{array}{*{20}{c}}
{\cos {\theta _i}}&{ - \sin {\theta _i}}\\
{\sin {\theta _i}}&{\cos {\theta _i}}
\end{array}} \right]^{\rm{T}}}\,,
\end{array}
	\label{eq3}
\end{equation}
Assuming that the dynamic obstacle can be wrapped by a minimum outer circle with radius $r_i^o$, the parameters ${a_i} = r_i^o + {v}_i^ot/2$ and ${b_i} = r_i^o$ are defined for the inner elliptical circle in Fig. \ref{fig3}. According to the above formulas, the probability field for the moving obstacle is established, and its partial derivative concerning the position can be calculated:

\begin{equation}
	\frac{{\partial U\left( {{\bf{p}}_i^o,{\bf{v}}_i^o,{\bf{p}}} \right)}}{{\partial {\bf{p}}}} =  - U{{\bm{\Sigma }}^{ - 1}}({\bf{p}} - {\bf{p}}_{{c_i}}^o)\,.
	\label{eq4}
\end{equation}

\subsection{Multi-robot Trajectory Generation}
In complex dynamic environments with uncertainties, the rapid generation of safe trajectories is essential for robots to respond to environmental changes on time and navigate safely. This section generates a safe trajectory in an uncertain environment using an unconstrained optimization method based on the safe probability field.
\subsubsection{Trajectory representation}
For the following system:
\begin{equation}
{\dot{\bf{x}}} = f({\bf{x}},{\bf{u}})\,,
\label{eq5}
\end{equation}
where $f$ is the state transition function, ${\bf{x}}$ is the state variable, and ${\bf{u}}$ is the input. The system is said to be differentially flat \cite{fliess1995flatness} if ${\bf{x}}$ and ${\bf{u}}$ can be represented by an analytic expression with a flat output ${\bf{z}}$ and its finite derivatives. In this case, only the higher-order differentiable trajectory of the center of mass and its yaw angle need to be planned for the trajectory planning of the system.

According to the differential flatness property, the MINCO (Minimum Control) trajectory \cite{wang2022geometrically} parameterization technology can simplify high-dimensional trajectories into a series of waypoints $\left\{ {{\bf{p}}_{11}^r, \ldots,{\bf{p}}_{1k}^r,\ldots,{\bf{p}}_{ij}^r,\ldots,{\bf{p}}_{Nk}^r,{\bf{p}}_e^r}\right\}$ and corresponding time $t_{ij}$ to flexibly respond to complex and changing environments, where ${\bf{p}}_{11}^r$ and ${\bf{p}}_e^r$ represent the starting position and ending position, respectively. The notation ${\bf{p}}_{ij}^r$ denotes the $j$th waypoint of the $i$th segment within the trajectory. The variable $N$ is the total number of segments, while $k$ indicates the number of waypoints per segment. For notational convenience, this paper adopts the shorthand ${\bf{p}}_{ij}^r \equiv {\bf{p}}_{(i - 1)k + j}^r$. Consequently, the set of waypoints can be represented as:
\begin{equation}
	S = \left\{ {{\bf{p}}_1^r, \ldots ,{\bf{p}}_i^r, \ldots ,{\bf{p}}_{Nk}^r,{\bf{p}}_{Nk + 1}^r} \right\},{\bf{p}}_i^r \in \mathbb{R}^n\,,
	\label{eq6}
\end{equation}
where $n$ is the dimension of the motion space. For the cost function $G({\bf{P}},{\bf{T}}) = H({\bf{C}}({\bf{P}},{\bf{T}}),{\bf{T}})$, $G$ and $H$ are the same costs in two different forms. The partial derivatives of the cost concerning the waypoints ${{\bf{p}}_i}$ and the unconstrained time ${T_i}$ are derived from the subsequent equations:
\begin{equation}
\begin{aligned}
\frac{{\partial G}}{{\partial {{\bf{p}}_i}}} &= {\mathop{\rm \mathrm{tr}}\nolimits} \left\{ {{{\left( {\frac{{\partial {\bf{C}}}}{{\partial {{\bf{p}}_i}}}} \right)}^{\rm{T}}}\frac{{\partial H}}{{\partial {\bf{C}}}}} \right\}\\
 &= {\mathop{\rm \mathrm{tr}}\nolimits} \left\{ {{{\left( {{{\bf{L}}^{ - 1}}\frac{{\partial {\bf{\left(LC\right)}}}}{{\partial {{\bf{p}}_i}}}} \right)}^{\rm{T}}}\frac{{\partial H}}{{\partial {\bf{C}}}}} \right\}\,,
 \label{eq7}
\end{aligned}
\end{equation}

\begin{equation}
\begin{aligned}
\frac{{\partial G}}{{\partial {T_i}}} &= \frac{{\partial H}}{{\partial {T_i}}} + {\mathop{\rm \mathrm{tr}}\nolimits} \left\{ {{{\left( {\frac{{\partial {\bf{C}}}}{{\partial {T_i}}}} \right)}^{\rm{T}}}\frac{{\partial H}}{{\partial {\bf{C}}}}} \right\}\\
 &= \frac{{\partial H}}{{\partial {T_i}}} - {\mathop{\rm \mathrm{tr}}\nolimits} \left\{ {{{\left( {\frac{{\partial {\bf{L}}}}{{\partial {T_i}}}{\bf{C}}} \right)}^{\rm{T}}}{{\bf{L}}^{ - {\rm{T}}}}\frac{{\partial H}}{{\partial {\bf{C}}}}} \right\}\,,
 \label{eq8}
\end{aligned}
\end{equation}
where ${\bf{C}}$ is the coefficient matrix, $\bf{L}$ is the condition matrix linking the trajectory segments, and ${\rm{tr}}\left\{\cdot\right\}$ denotes the operation of obtaining the trace of the matrix. The specific settings can be related to the work\cite{wang2022geometrically}. 

\subsubsection{Optimization problem construction}
According to the Eq.\eqref{eq14} and Eq.\eqref{eq15}, the omnidirectional mobile robot studied in this paper has a differential flatness property similar to the quadrotor when the flat output ${\bf{z}} = \left[ {{x\rm{_w}},{y\rm{_w}},{\theta \rm{_{bw}}}} \right]$ is set. Combining this differential flatness with the MINCO trajectory parameterization method transforms the trajectory optimization problem into an unconstrained problem to speed up the optimization process. The overall cost function is defined as:
\begin{equation}
	\begin{aligned}
		{C_{{\rm{tot }}}} = {C_{{\rm{ego}}}} +  {\omega _{{\rm{dyn}}}} \cdot {C_{{\rm{dyn}}}}\,,
		\label{eq 9}
	\end{aligned}
\end{equation}
where ${C_{{\rm{ego}}}}$ is a weighted combination of the smoothness cost, dynamic feasibility cost, time cost, swarm avoidance cost, and static obstacle avoidance cost. ${C_{{\rm{dyn}}}}$ and ${\omega _{{\rm{dyn}}}}$ are the cost and weight of dynamic uncertain obstacle avoidance, respectively. This paper discusses the details of the dynamic obstacle avoidance component and refers to previous work \cite{zhou2022swarm} for the ${C_{{\rm{ego}}}}$. 

\begin{figure}[htbp]
	\centering
	\includegraphics[width = 1.0\linewidth, keepaspectratio]{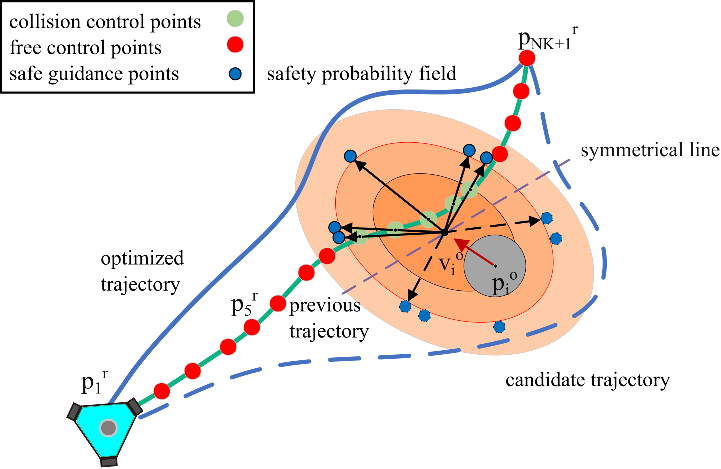}
	\caption{Illustration of trajectory optimization based on safety probability field. For the trajectory that detects collision risks (the green curve), the collision probability values of dangerous path points (the light green dot) and the gradient (the black solid line) away from collisions can be obtained based on the safety probability field. Subsequently, the current dangerous path point ${\bf{p}}_i^r$ is to be advanced to the safe guidance point ${\bf{p}}_{ij}^g$ (the blue dot) along the gradient. Furthermore, by establishing a symmetrical line (the purple dashed line) through the index of path points entering and exiting the risk area, it is possible to obtain a candidate trajectory (the blue dashed line) on the other side of the dynamic obstacle.}
	\label{fig4}
\end{figure}

For the collision cost and gradient required for optimization, the safety probability field is employed to find guidance points for estimating the collision cost and gradient of dynamic obstacles with uncertainty. The process of setting specific guidance points is illustrated in Fig. \ref{fig4}. SPF can calculate the probability of a collision occurring at a given point. A collision probability exceeding the prescribed threshold can indicate a potential collision. When a collision risk is detected for the given waypoint within the current trajectory, a guide point is initially sought according to the safe gradient of the waypoint. After that, a collision-free trajectory can be generated. Concurrently, to ensure the completeness of the trajectory solution, an alternative trajectory is obtained on the other side according to the symmetry line. Ultimately, the optimal trajectory is selected based on a comparison of the trajectory costs. It is worth noting that optimization-based methods are inherently susceptible to oscillation due to the dynamic nature of the environment. However, engineering settings can improve this limitation, for example, by automatically adjusting the threshold based on the number of oscillations.

According to the SPF, each dangerous waypoint ${\bf{p}}_i^r$ for dynamic obstacles ${\bf{p}}_j^o$ is paired with a corresponding guidance point ${\bf{p}}_{ij}^g$. Subsequently, using the predefined safety threshold ${d_{{\rm{saf }}}}$, the collision cost and gradient at the collision control point can be calculated as follows:
\begin{equation}
	{C_{{\rm{dyn }}}} = \sum\limits_i {\sum\limits_j {{\lambda _i}{{\left( {{d_{{\rm{saf}}}} - \left( {{\bf{p}}_i^r - {\bf{p}}_{ij}^g} \right) \cdot \frac{{\nabla {{\bf{g}}_{ij}}}}{{\left\| {\nabla {{\bf{g}}_{ij}}} \right\|}}} \right)}^3}} }\,,
	\label{eq10}
\end{equation}
\begin{equation}
	\nabla {{\bf{g}}_{ij}} = U\left( {{\bf{p}}_j^o,{{\bf{v}}_j},{\bf{p}}_i^r} \right){{\bm{\Sigma }}^{ - 1}}({\bf{p}}_i^r - {\bf{p}}_{{c_j}}^o)\,,
	\label{eq11}
\end{equation}
where $i$ and $j$ denote the path point and obstacle serial numbers, respectively, the weight assigned to the path point is ${\lambda _i}$. The negative gradient direction of the SPF, denoted by $ \nabla {{\bf{g}}_{ij}}$, can influence the trajectory by guiding waypoints towards a safe region.

The gradient of the collision cost concerning the trajectory polynomial coefficients $\bf{C}$ and time $\bf{T}$ are calculated using the chain rule: 
\begin{equation}
	\begin{aligned}
		\frac{{\partial {C_{\rm{dyn}}}}}{{\partial \bf{C}}} = & - 3\sum\limits_i {\sum\limits_j {{\lambda _i}} } {\bf{\beta }}\left( {{t_i}} \right){\left( {\frac{{\nabla {{\bf{g}}_{ij}}}}{{\left\| {\nabla {{\bf{g}}_{ij}}} \right\|}}} \right)^{\rm{T}}}&\\
		&\cdot{\left( {{d_{{\rm{saf}}}} - \left( {{\bf{p}}_i^r - {\bf{p}}_{ij}^g} \right) \frac{{\nabla {{\bf{g}}_{ij}}}}{{\left\| {\nabla {{\bf{g}}_{ij}}} \right\|}}} \right)^2}\,,&
		\label{eq12}
	\end{aligned}
\end{equation}

\begin{equation}
	\begin{aligned}
		\frac{{\partial {C_{\rm{dyn}}}}}{{\partial \bf{T}}} =  &- 3\sum\limits_i {\sum\limits_j {{\lambda _i}} } {{\bf{v}}_i}{\bf{\alpha }}_i{\left( {\frac{{\nabla {{\bf{g}}_{ij}}}}{{\left\| {\nabla {{\bf{g}}_{ij}}} \right\|}}} \right)^{\rm{T}}}&\\
		&\cdot{\left( {{d_{{\rm{saf}}}} - \left( {{\bf{p}}_i^r - {\bf{p}}_{ij}^g} \right) \frac{{\nabla {{\bf{g}}_{ij}}}}{{\left\| {\nabla {{\bf{g}}_{ij}}} \right\|}}} \right)^2}\,,&
		\label{eq13}
	\end{aligned}
\end{equation}
where ${\bf{\beta}}\left( {{t_i}} \right) = \left[ 1,t_i,{t_i}^2,{t_i}^3,{t_i}^4,{t_i}^5\right]$ is the basis function of the trajectory segment, ${\bf{\alpha }}_i = i/k$ is the scaling factor of the trajectory segment, and ${\bf{v}}_i$ is the required speed corresponding to the waypoint. According to Eq.\eqref{eq7} and \eqref{eq8}, the derivatives of the dynamic collision cost concerning the waypoints and the unconstrained time can be calculated. After that, an unconstrained optimization accelerates the trajectory generation process.

\subsection{Trajectory tracking Control}

In practice, internal and external disturbances introduce uncertainty when observing the robot's state. The uncertainty can lead to trajectory tracking errors, thereby diminishing the efficacy of the generated trajectories. Consequently, a robust controller that accounts for such uncertainty is essential for safe navigation. In this section, an extended state observer is designed for the uncertain state of the robot, and the iterative trajectory is tracked in combination with self-sampling model predictive control.

\subsubsection{Robot model}
\begin{figure}[htbp]
	\centering
        \includegraphics[width = 0.8\linewidth, keepaspectratio]
        {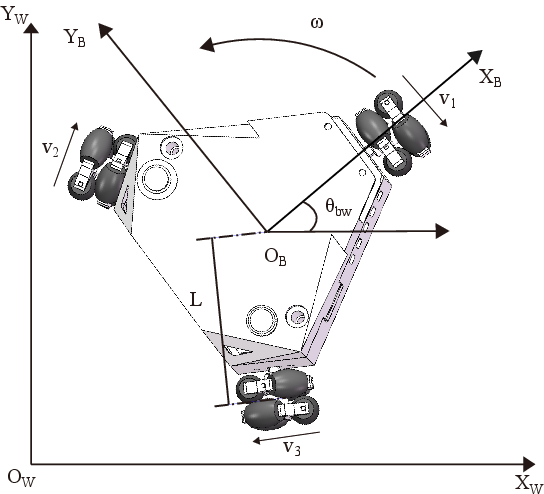}
	\caption{Robot model and relative relationship between the coordinate systems, where ${\rm{L=0.04m}}$, ${\theta _{\rm{bw}}}$ is the angle between the corresponding coordinate systems. The positive directions of the velocities are given.}
	\label{fig5}
\end{figure}

The three-wheeled omnidirectional mobile robot employed in this paper is characterized by its flexible motion capabilities, as depicted in Fig. \ref{fig5}.

The positive direction of the robot body coordinate system's x-axis corresponds to the wheel axis's direction, and the wheels are evenly distributed at intervals of $120^{\circ}$. The kinematic and kinetic equations of the robot are as follows:

\begin{equation}
	\mathop {\dot{\bf{z}}} = \left[ {\begin{array}{*{20}{c}}
{{v_{wx}}}\\
{{v_{wy}}}\\
\omega 
\end{array}} \right] = {\bf{A}}{\left[ {\begin{array}{*{20}{c}}
0&{ - 1}&{{\rm{ - L}}}\\
{\cos {{60}^\circ }}&{\sin {{60}^\circ }}&{{\rm{ - L}}}\\
{ - \cos {{60}^\circ }}&{\sin {{60}^\circ }}&{{\rm{ - L}}}
\end{array}} \right]^{ - 1}}\left[ {\begin{array}{*{20}{c}}
{{v_1}}\\
{{v_2}}\\
{{v_3}}
\end{array}} \right]+{\bf w}_v\,,
	\label{eq14}
\end{equation}

\begin{equation}
	{\ddot{\bf z}} = \left[ {\begin{array}{*{20}{c}}
{{{\dot v}_{wx}}}\\
{{{\dot v}_{wy}}}\\
{\dot \omega }
\end{array}} \right] = {\bf{A}}\left[ {\begin{array}{*{20}{c}}
{ - \frac{{\sqrt 3 }}{{2{\rm{MR}}}}}&{ - \frac{{\sqrt 3 }}{{2{\rm{MR}}}}}&0\\
{ - \frac{1}{{2{\rm{MR}}}}}&{ - \frac{1}{{2{\rm{MR}}}}}&{\frac{1}{{{\rm{MR}}}}}\\
{ - \frac{{\rm{L}}}{{{\rm{IR}}}}}&{ - \frac{{\rm{L}}}{{{\rm{IR}}}}}&{ - \frac{{\rm{L}}}{{{\rm{IR}}}}}
\end{array}} \right]\left[ {\begin{array}{*{20}{c}}
{{\tau _1}}\\
{{\tau _2}}\\
{{\tau _3}}
\end{array}} \right]+{\bf w}_a\,,
	\label{eq15}
\end{equation}

\begin{equation}
	\mathop {\bf{A}}\limits = \left[ {\begin{array}{*{20}{c}}
{\cos {\theta _{{\rm{bw}}}}}&{ - \sin {\theta _{{\rm{bw}}}}}&0\\
{\sin {\theta _{{\rm{bw}}}}}&{\cos {\theta _{{\rm{bw}}}}}&0\\
0&0&1
\end{array}} \right]\,,
	\label{eq16}
\end{equation}
where ${\bf{A}}$ is the coordinate transformation matrix. ${v_1}$, ${v_2}$, and ${v_3}$ represent the linear velocities of the wheels, ${\tau _1}$, ${\tau _2}$, and ${\tau _3}$ represent the control torque of each wheel, and $\omega$ is the rotational angular velocity of the robot. The ${v_{bx}}$ and ${v_{wx}}$ are the velocities of the robot in its body coordinate system and world coordinate system, respectively. ${\rm{M}}$ is the mass of the robot, ${\rm{R}}$ is the radius of the wheel, and ${\rm{I}}$ is the moment of inertia. ${\bf w}_v$ and ${\bf w}_a$ are the observation errors for the kinematic and kinetic states, respectively.

\subsubsection{Extended state MPC}
Uncertainties arising from model inaccuracies, sensor noise, and system delays can be collectively attributed to the uncertainty in the observed state. Consequently, an Extended State Observer (ESO) \cite{han2009pid} is employed to monitor the expanded perturbed state and provide compensation for the actual observed state:
\begin{equation}
\left[ {\begin{matrix}
{{{\dot {\hat z}}}}\\
{{{\ddot {\hat z}}}}
\end{matrix}} \right] = \left[ {\begin{matrix}
{ - {l_1}}&1\\
{ - {l_2}}&0
\end{matrix}} \right]\left[ {\begin{matrix}
{{{\hat z}}}\\
{\dot{\hat z}}
\end{matrix}} \right] + \left[ {\begin{matrix}
b&{{l_1}}\\
0&{{l_2}}
\end{matrix}} \right]\left[ {\begin{matrix}
u\\
z
\end{matrix}} \right]\,,
\label{eq17}
\end{equation}
where $x$ is the measured state variable, $\hat{x}$ is the state estimation, and $u$ is the control input. In the kinematic model, the control gain is set to $b = 1$, and the ESO gains are $l_1 = 2w$ and $l_2 = w^2$, with the poles of the system uniformly configured at $-w$.

For local trajectories that are continuously iterated and optimized by the planner, this work fits the polynomial trajectory using the positions, velocities, and timestamps of adjacent waypoints in the reference trajectory, effectively transforming waypoint data into trajectory parameters:
\begin{equation}
	{{\bf traj}_{{\rm{ref, }}k}} = {\bf traj} \left( {{{\hat{\bf{z}}}_0}},\,{{t_0} + (k - 1)dt} \right),\,\forall k = 1,2, \ldots ,H\,,
	\label{eq18}
\end{equation}
where ${{\hat{\bf{z}}}_0}$ is the current state after ESO processing. Adaptive sampling is executed based on the current time $t_0$, the horizon of time $H$, the sampling interval $dt$, and the trajectory parameter information. Finally, the optimal control problem is formulated as follows:
\begin{subequations}
	\begin{align}
		\mathop {\min }\limits_{{{\bf{z}}_k},{{\bf{u}}_k}}
		\quad&\sum\limits_{k = 0}^{H - 1} {\left( {{\lambda _p}{{\left\| {{{\bf{p}}_k} - {{\bf traj}_{{\rm{ref, }}k}}} \right\|}^2} + {\lambda _u}{{\left\| {{{\bf{u}}_k}} \right\|}^2}} \right)}&\\
		\mathrm{s.t.} \quad & {{\hat{\bf{z}}}_{k}} = {\rm{eso}} \left({{\bf{z}}_k, {\bf{u}}_k}\right) \,,& \\
  & {{\bf{z}}_{k + 1}} = f\left( {{{\hat{\bf{z}}}_{k}},{{\bf{u}}_k}} \right) + {\bf{w}}_k \,,& \\
		&{{\bf{z}}_0} = {{\bf{z}}_{{\rm{init}}}}\,,&  \\
		&{{\bf{z}}_{{\rm{lb}}}} \le {{\bf{z}}_k} \le {{\bf{z}}_{{\rm{ub}}}} \,,& \\
		&{{\bf{u}}_{{\rm{lb}}}} \le {{\bf{u}}_k} \le {{\bf{u}}_{{\rm{ub}}}} \,, &
		\label{eq19}
	\end{align}
\end{subequations}
where the state variables are represented by the vector ${\bf{z}} = \left[ {{x\rm{_w}},{y\rm{_w}},{\theta \rm{_{bw}}}} \right]$, and the control inputs are represented by the vector ${\bf{u}} = \left[ {{v_1},{v_2},{v_3}} \right]$. The ${\bf{w}}_k$ denotes the total process noise. The non-zero track error factor ${\lambda_p}$ and control input factor ${\lambda_u}$ are designed to limit the solution space to balance tracking accuracy and control input smoothness.

\section{Results and Discussion}\label{sec3}

To further assess the efficacy of the proposed method, we perform a series of experiments in both dynamic simulation and real-world scenarios. The L-BFGS algorithm \cite{liu1989limited} and the CasADi toolkit \cite{andersson2019casadi} are utilized to address the unconstrained trajectory optimization and the optimal control problems, respectively. These algorithms are integrated into the ROS environment through C++ code. Separate ROS nodes have been engineered to allow for a transition between simulation and physical application without altering the codebase, operating in a distributed manner on an Intel i9-13900K. The full experimental video is available at \url{https://www.bilibili.com/video/BV1xpmNYJE5F/}.

\begin{figure}[htbp]
	\centering
	\includegraphics[width = 1.0\linewidth, keepaspectratio]{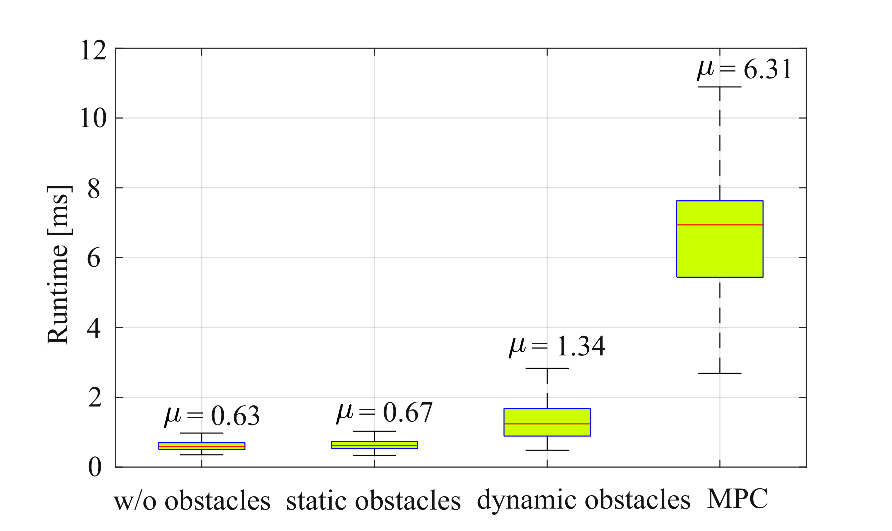}
	\caption{The proposed planner and controller runtime under various conditions. The computational time for optimization slightly increases when the robot encounters dynamic obstacles, but real-time performance can still be ensured. The controller runs simultaneously with the planner and supports multiple planning by the planner.}
	\label{fig6}
\end{figure}

Fig. \ref{fig6} illustrates the execution time for each algorithm segment. In the trajectory generation planner, the construction of the SPF requires a certain amount of runtime when faced with dynamic obstacles. However, this does not affect the planner's real-time performance. The total duration of the iterations is less than 10 ms, allowing for a timely response to dynamic changes in the environment.

\subsection{Experimental Setup}
For trajectory planning, we set the attention in Eq.\eqref{eq2} $\lambda=0.5$ in the safe probability field and select a fifth-order polynomial for trajectory parameterization in the unconstrained optimization, achieving a balance between the algorithm's real-time performance and trajectory smoothness. In trajectory tracking, the ESO gains in Eq.\eqref{eq17} are set to $l_1 = 1$ and $l_2 = 0.25$. The model predictive control's time domain in Eq.\eqref{eq18} is configured with $H=5$, using a time step of $dt=0.1s$. 

To address environmental uncertainties, we assume that the state of the $i$th obstacle at the next time step conforms to ${\bf{p}}_{i,{t_{k + 1}}}^o = {\bf{p}}_{i,{t_k}}^o + {\bf{v}}_{i,{t_k}}^{o'}\left( {{ t_{k + 1}} - {t_k }} \right) + {\bf{p}}_{ie}^o$ and ${\bf{v}}_{i,{t_k}}^{o'} = {\bf{v}}_{i,{t_k}}^o + {\bf{v}}_{ie}^o$. Both the position deviation ${\bf{p}}_{ie}^o$ and the velocity deviation ${\bf{v}}_{ie}^o$ of the obstacles adhere to a Gaussian distribution. Furthermore, the state deviations (1m deviation between the x and y states, $0.1^\circ$ deviation in $\theta$) are introduced for the robot in the simulation, thereby simulating the uncertainty of the observed state.

\subsection{Simulation Experiment}
\begin{figure*}[!ht]
	\centering
	\includegraphics[width=18cm, keepaspectratio]{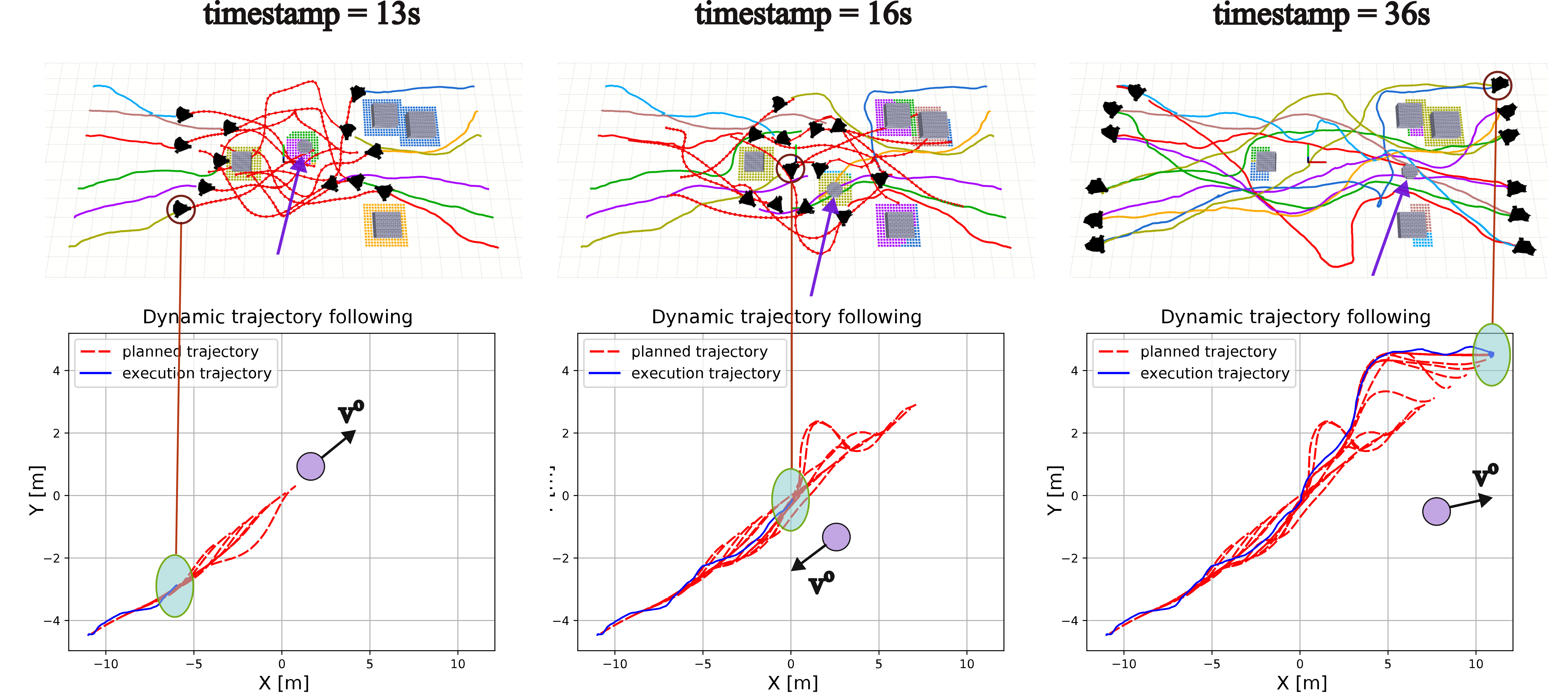}
	\caption{Demonstration of multi-robot navigation at different moments in a cluttered environment. The upper figure shows an example of 12 robots safely navigating in a simulation scenario containing static obstacles (the square) and an uncertain dynamic obstacle (the circle indicated by a purple arrow), and different colors under the obstacles denote the perceptions of different robots. The lower figure shows the circled robot using EMPC to track iterative local trajectories under the uncertainty indicated by the light green area, where the state of the moving obstacle (the purple circle) is also shown.}
\label{fig7}
\end{figure*}

A multi-robot simulation is conducted in a complex dynamic environment, as depicted in Fig. \ref{fig7}. Each robot is capable of independently avoiding obstacles and safely reaching its destination. Their target positions are set in opposing configurations to ensure comprehensive interaction among the robots.

To verify the effectiveness of the algorithm, this paper compares the trajectory generation aspect with the current state-of-the-art method and conducts ablation studies on individual components. These assessments aim to ascertain the necessity and superiority of the proposed method.

\subsubsection{Comparison of trajectory generation}
\begin{table*}[t]
    \caption{Comparison with EGO-Swarm2 (ES2), its constant prediction algorithm (ES2*), and our SPF-EMPC method. Fifty simulations at varying uncertainty levels (1\,${\Sigma}: {\sigma_{p}}^o = 1\times{10^{-4}}, {\sigma_{v}}^o = 1\times{10^{-2}}$) were conducted in a randomly generated environment similar to Fig. \ref{fig7}, which included five robots, four dynamic obstacles, and six static obstacles. The metrics are the success rate (SR), overtime rate (OR), dynamic distance, path length, travel time, and iterations. Each metric's mean $|$ minimum $|$ standard deviation are presented. The better performer among methods is highlighted in bold.}
	\centering
	\begin{tabular}{cccccccc}
		\hline
		Uncertainty Level    & Methods                  & SR (\%) & OR (\%) & Dynamic Distance (m) & Path Length (m) & Travel Time (s) & Iterations \\ \hline
		\multirow{3}{*}{0.25\,${\Sigma}$} & ES2              & 14  & \textbf{0} & 2.30 $|$ 0.67 $|$ 1.70                 & \textbf{25.43} $|$ \textbf{25.13} $|$ 0.44           & \textbf{34.22} $|$ 32.00 $|$ 4.15           & 253.74     \\ 
		& ES2* & \textbf{76}                & 10                 & 1.76 $|$ 0.97 $|$ 1.11                & 26.41 $|$ 25.15 $|$ 1.29          & 35.30 $|$ \textbf{31.01} $|$ 2.71          & 69.92      \\
		& Ours                 & \underline{72}                & 2                  & \textbf{1.52} $|$ \textbf{0.62} $|$ 0.85                & 27.17 $|$ 25.14 $|$ 2.42          & 38.83 $|$ 32.00 $|$ 7.02          & \textbf{58.4}       \\ \hline
		\multirow{3}{*}{1\,${\Sigma}$}    & ES2                 & 10                & \textbf{0}                  & 1.50 $|$ 0.73 $|$ 0.79  & \textbf{25.62} $|$ \textbf{25.15} $|$ 0.44           & \textbf{31.80} $|$ \textbf{31.00} $|$ 0.57           & 270.16     \\
		& ES2* & \textbf{62}                & 4                  & 1.41 $|$ 0.85 $|$ 3.41                 & 26.53 $|$ 25.28 $|$ 1.33          & 36.08 $|$ 31.50 $|$ 3.59          & 92.52      \\
		& Ours                 & \underline{56}                & 2                  & \textbf{1.38} $|$ \textbf{0.69} $|$ 0.62                 & 29.64 $|$ 25.18 $|$ 4.44           & 39.34 $|$ 33.50 $|$ 5.10          & \textbf{55.78}      \\ \hline
		\multirow{3}{*}{2\,$\Sigma$}    & ES2                & 12                & 2                  & \textbf{1.39} $|$ 0.57 $|$ 0.68                & \textbf{26.21} $|$ 25.20 $|$ 1.28          & \textbf{33.33} $|$ \textbf{31.50} $|$ 1.57          & 300.96     \\
		& ES2* & \textbf{52}                & 4                 & \textbf{1.39} $|$ 0.57 $|$ 0.68               & 27.37 $|$ 25.46 $|$ 1.88        & 38.33 $|$ 33.00 $|$ 5.37       & 101.74     \\
		& Ours                 & \underline{40}                 & \textbf{2}                  & 1.47 $|$ \textbf{0.53} $|$ 1.01             & 29.35 $|$ \textbf{25.17} $|$ 3.71       & 39.60 $|$ \textbf{31.50} $|$ 5.35       & \textbf{60.36}   \\ \hline  
	\end{tabular}
    \label{tab2}
\end{table*}

The local planner EGO-Swarm2 \cite{zhou2022swarm} and its variant with a constant prediction for dynamic obstacles serve as benchmarks. The comparative data at varying levels of uncertainty are presented in Table \ref{tab2}. Among the performance metrics, the overtime rate indicates the likelihood that the robot is trapped in the environment, failing to reach its target. The number of iterations refers to the mean number of optimizations required for a single robot within a single experimental trial.

As illustrated in Table \ref{tab2}, the success rate of the three methods declines gradually with increasing uncertainty. Notably, ES2 exhibits the lowest overtime rate and requires the shortest distance and time to navigate. However, ES2 perceives dynamic obstacles as static in real-time and does not account for uncertainty, resulting in the lowest success rate and most iterations. The ES2* variant, which assumes obstacles follow perfectly predictable trajectories, markedly enhances the success rate and iteration count compared to ES2. Nevertheless, it tends to induce oscillations when confronted with uncertain obstacles, leading to a higher overtime rate. Moreover, its ideal prediction model is challenging to implement in practical scenarios. In contrast, the proposed method achieves a considerably higher success rate than ES2 and nearly matches the ideal ES2* performance, with the fewest iterations and dynamic distance. The result indicates that the SPF-based unconstrained planning approach converges rapidly and avoids excessive conservatism in generating safe trajectories.

\subsubsection{Ablation comparison}
\begin{table}[htb]
\caption{Performance comparison of the proposed SPF-EMPC with other versions at different uncertainty levels ($1\,{\Sigma'}$: 1m deviation between the x and y states, $0.1^\circ$ deviation in $\theta$) in a simulated environment, based on success rate (SR), overtime rate (OR), and travel time (TT).}
\centering
\begin{tabular}{ccccc}
\hline
Uncertainty Level & Methods      & SR (\%)    & OR (\%)    & TT (s)     \\ \hline
\multirow{3}{*}{$0\,\Sigma'$}  & SPF-EMPC w/o SS  & 48          & \textbf{0}          & 47.50          \\
                    & SPF-EMPC w/o SPF & 56          & 4 & \textbf{30.00}          \\
                    & Ours     & \textbf{86} & 2          & 33.12 \\ \hline
\multirow{2}{*}{$1\,\Sigma'$}  & SPF-EMPC w/o ESO & 42          & 2          & 76.43          \\
                    & Ours     & \textbf{88}          & \textbf{0}          & \textbf{33.42}   \\ \hline
\end{tabular}
\label{tab3}
\end{table}

To substantiate the efficacy of each component of our proposed method, we performed an ablation comparison, as shown in Table \ref{tab3}. Specifically, in the variant lacking self-sampling (SS), the trajectory referenced in Eq.\eqref{eq18} is substituted with a generic MPC based on the initial discrete trajectory. For the version without SPF, the speed of the safety probability field is set to zero, degenerating into a circular field with the minimum safe distance.

The data in the table shows that even under ideal observation conditions (${0\,\Sigma'}$), the SPF-EMPC cannot accurately track the iterative trajectory without SS, ultimately resulting in the lowest recorded success rate. Furthermore, the absence of the SPF severely affects the system's ability to avoid dynamic obstacles, resulting in a significant decrease in success rate. Considering the inherent uncertainty in the robot's state, the lack of ESO complicates the controller's ability to accurately monitor the state, potentially issuing erroneous commands and increasing the complexity of real-time optimization. As a result, this leads to a decrease in success rate and a significant increase in travel time. These findings demonstrate the effectiveness and necessity of the integrated components in our proposed approach, especially when facing uncertain obstacles and the robot's imprecise state.

\begin{figure}[!ht]
	\centering
	\subfigure[]{\includegraphics[width = 1.0\linewidth, keepaspectratio]{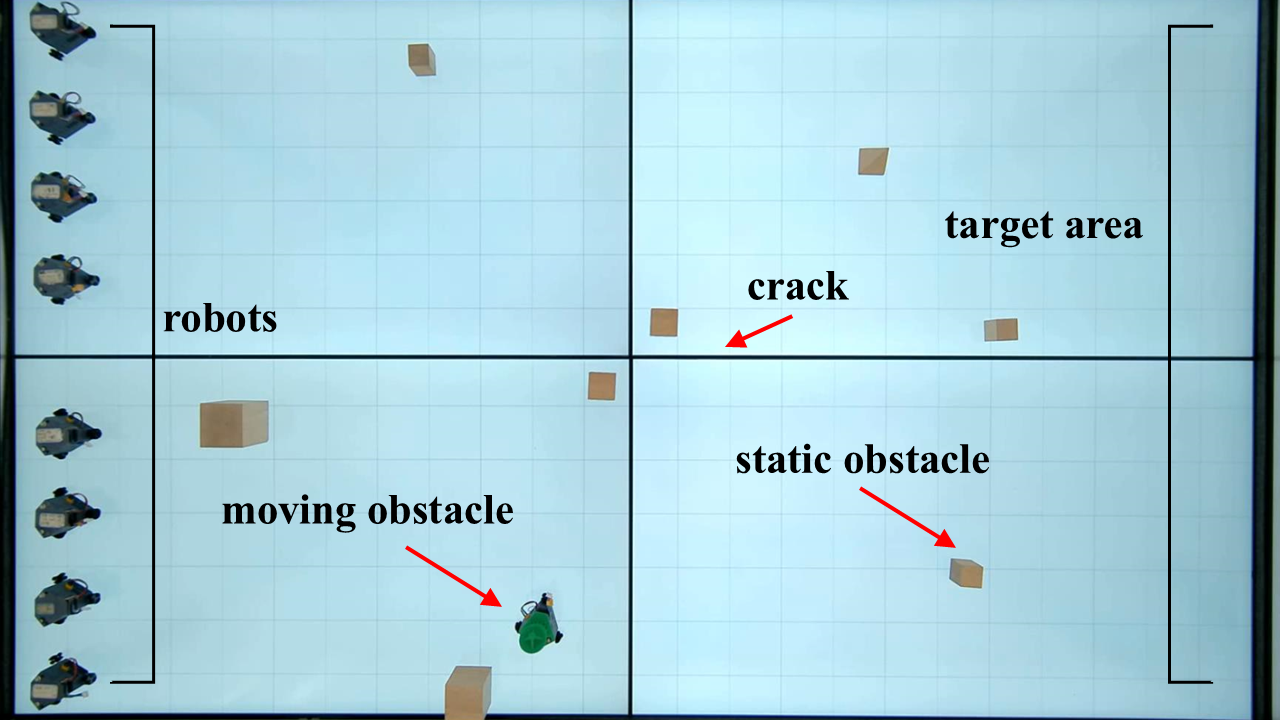}}
	\subfigure[]{\includegraphics[width = 1.0\linewidth, keepaspectratio]{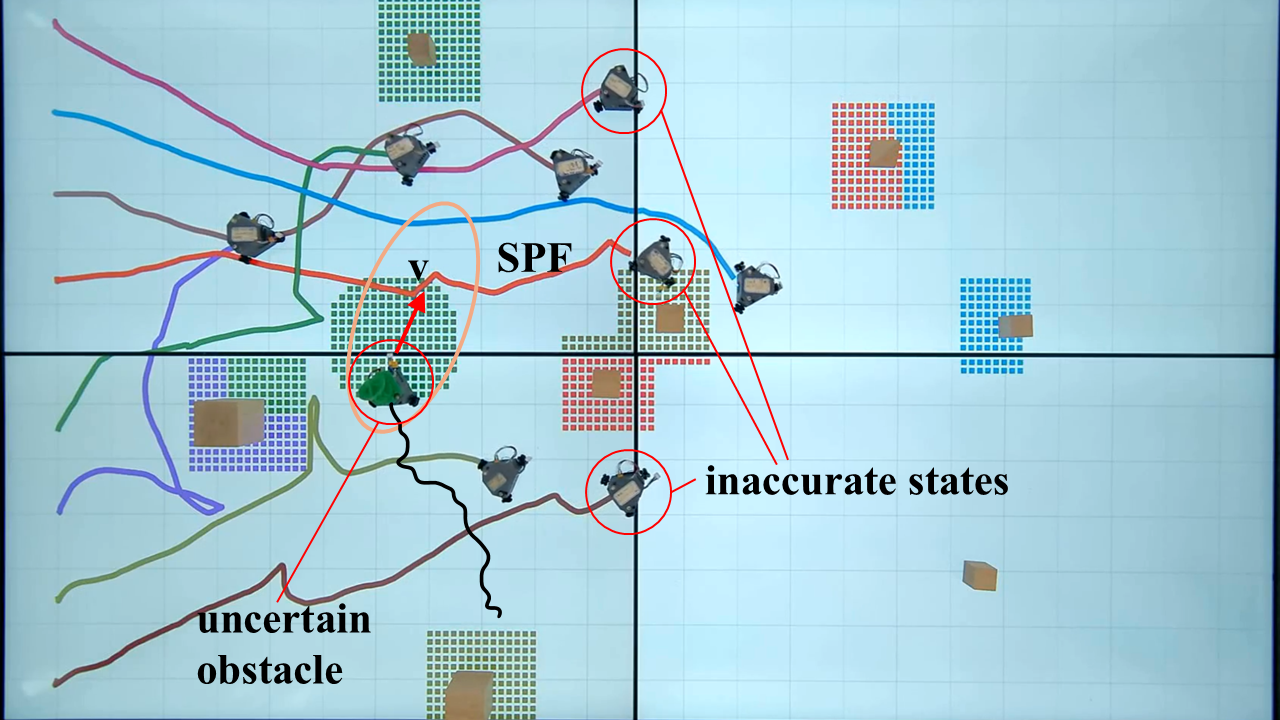}}
        \subfigure[]
     {\includegraphics[width = 1.0\linewidth, keepaspectratio]{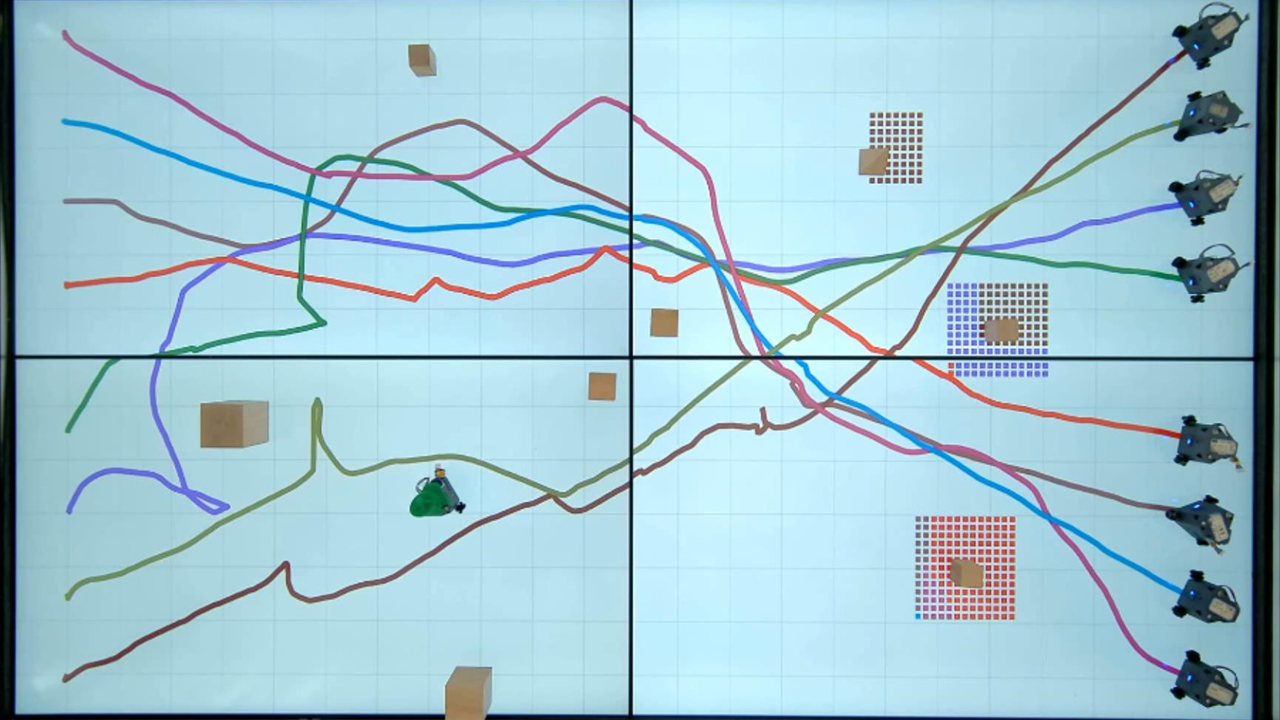}}
	\caption{Illustration of a physical navigation scenario within the dynamic environment. (a) An introduction to the physical environment, which includes eight robots, eight static obstacles, and one dynamic obstacle. (b) The real-time response process to a changing environment that includes uncertainties caused by the random movement of an obstacle and imprecise observation of the robot's states at cracks. (c) All robots reach their destination efficiently without any collisions.}
	\label{fig8}
\end{figure}

\subsection{Physical Experiment}

In a real-world scenario, a scaled-down platform \cite{zhu2024dvrp}, measuring $1.374 \times 2.432 \text{m}$, was constructed for conducting experiments. A three-wheeled omnidirectional robot was designed for physical testing, as shown in Fig. \ref{fig1}.

Fig. \ref{fig8} shows an example of actual running. In the experiment, several static obstacles and one dynamic obstacle are set up to hinder the movement of the robots. The motion of the dynamic obstacle, the interference at the platform seams, and the robots' traveling all introduce uncertainties. Robots need to utilize their limited perception and communication capabilities to avoid various obstacles and other robots, and ultimately reach the goal point. Practical experiments demonstrate that the proposed method can enable the safe navigation of multiple robots in a dynamic environment with uncertainties.

\begin{figure}[htbp]
	\centering
	\includegraphics[width=8cm, keepaspectratio]{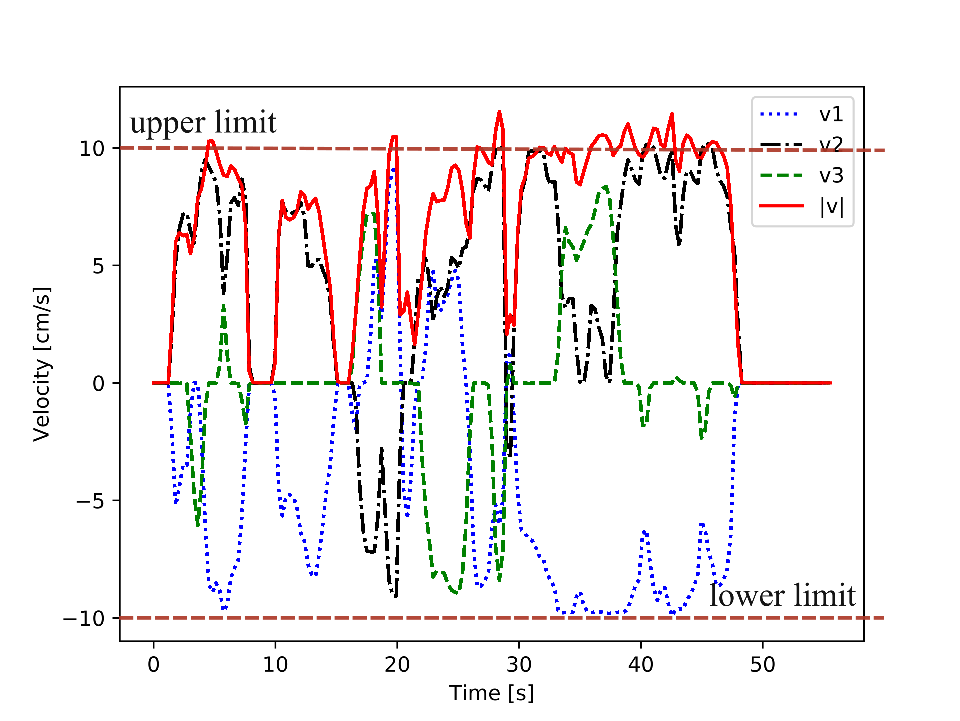}
	\caption{Wheel speed profile for each axis in the physical running test. The speed data are obtained via onboard motor state feedback.}
	\label{fig9}
\end{figure}

In this process, the results of the velocities returned by the robot are shown in Fig. \ref{fig9}. The speed of each wheel is restricted within the driving limit, reflecting the effective role of the controller.

\section{Conclusion and Future Work}\label{sec4}

This paper introduces a safe probability field based extended state model predictive control planner (SPF-EMPC) for multi-robot navigation in uncertain environments. During trajectory planning, a safe probability field is designed to model dynamic obstacles, enabling the online generation of iterative trajectories via an unconstrained optimization approach. Moreover, to accurately track the continuously updated planning trajectories under the driving constraints, discrete trajectory points are fitted to polynomial parameters for MPC adaptive sampling, and an extended state observer is employed to handle the uncertainty of robot state observation. Simulation results validate the efficacy and indispensability of each component of the proposed method, showcasing its ability to match the performance of the ego-swarm2 algorithm under ideal conditions with fewer iterations. The practical application on eight robots demonstrates the algorithm's real-time capability and reliability. Future work will focus on scaling up the multi-robot system and integrating perception to better handle uncertainties.









\bibliography{reference.bib}





\end{document}